\documentclass[letterpaper]{article} 
\usepackage{aaai2026}  
\usepackage{times}  
\usepackage{helvet}  
\usepackage{courier}  
\usepackage[hyphens]{url}  
\usepackage{graphicx} 

\usepackage{booktabs}
\usepackage{amsmath}
\usepackage{amsfonts} 
\usepackage{colortbl}
\usepackage{subfigure}
\usepackage{multirow}
\usepackage{caption}
\usepackage{subcaption}
\usepackage{tabularx}
\usepackage[table]{xcolor}
\usepackage{mathtools}

\usepackage{natbib}  
\usepackage{caption} 
\usepackage{algorithm}
\usepackage{algorithmic}

\usepackage{newfloat}
\usepackage{listings}
\DeclareCaptionStyle{ruled}{labelfont=normalfont,labelsep=colon,strut=off} 
\floatstyle{ruled}
\newfloat{listing}{tb}{lst}{}
\floatname{listing}{Listing}
\title{DiCaP: Distribution-Calibrated Pseudo-labeling for Semi-Supervised Multi-Label Learning}
\author{
Bo~Han\textsuperscript{\rm 1},
Zhuoming~Li\textsuperscript{\rm 1},
Xiaoyu~Wang\textsuperscript{\rm 2},
Yaxin~Hou\textsuperscript{\rm 1},
Hui~Liu\textsuperscript{\rm 4},
Junhui~Hou\textsuperscript{\rm 5},
Yuheng~Jia\textsuperscript{\rm 1,3,4}\thanks{Corresponding author.}
}
\affiliations{
\textsuperscript{\rm 1}School of Computer Science and Engineering, Southeast University, Nanjing, China\\
\textsuperscript{\rm 2}School of Software Engineering, Southeast University, Nanjing, China\\
\textsuperscript{\rm 3}Key Laboratory of New Generation Artificial Intelligence Technology and Its \\Interdisciplinary Applications (Southeast University), Ministry of Education, China\\
\textsuperscript{\rm 4}School of Computing Information Sciences, Saint Francis University, Hong Kong, China\\
\textsuperscript{\rm 5}Department of Computer Science, City University of Hong Kong, Hong Kong, China\\
\{hanbo,~leezhuoming,~wang\_xy,~yaxin,~yhjia\}@seu.edu.cn, 
hliu99-c@my.cityu.edu.hk, jh.hou@cityu.edu.hk
}

\usepackage{bibentry}
\begin{document}

\maketitle

\begin{abstract}
Semi-supervised multi-label learning~(SSMLL) aims to address the challenge of limited labeled data in multi-label learning~(MLL) by leveraging unlabeled data to improve the model performance. While pseudo-labeling has become a dominant strategy in SSMLL, most existing methods assign \textbf{equal weights} to all pseudo-labels regardless of their quality, which can amplify the impact of noisy or uncertain predictions and degrade the overall performance. In this paper, we \textbf{theoretically} verify that the optimal weight for a pseudo-label should reflect its correctness likelihood. \textbf{Empirically}, we observe that on the same dataset, the correctness likelihood distribution of unlabeled data remains stable, even as the number of labeled training samples varies. Building on this insight, we propose \underline{\textbf{Di}}stribution-\underline{\textbf{Ca}}librated \underline{\textbf{P}}seudo-labeling~(\textbf{DiCaP}), a correctness-aware framework that estimates posterior precision to calibrate pseudo-label weights. We further introduce a dual-thresholding mechanism to separate confident and ambiguous regions: confident samples are pseudo-labeled and weighted accordingly, while ambiguous ones are explored by unsupervised contrastive learning. Experiments conducted on multiple benchmark datasets verify that our method achieves consistent improvements, surpassing state-of-the-art methods by up to \textbf{4.27\%}. 
\end{abstract}
\begin{links}
 \link{Code}{https://github.com/hb-studying/DiCaP}
\end{links}

\section{Introduction}

\indent Multi-label learning~(MLL) addresses scenarios where each instance may be associated with multiple labels, enabling richer and more comprehensive predictions across complex tasks~\cite{EmergingMLL}. It has been widely adopted in various applications, such as image annotation~\cite{app_image}, text categorization~\cite{app_text}, and facial expression recognition~\cite{app_face}. However, a major challenge in real-world scenarios is the high cost associated with obtaining precise annotations for every instance. To alleviate this burden, semi-supervised multi-label learning~(SSMLL) has gained considerable interest. Specifically, SSMLL aims to construct effective models by utilizing a small portion of fully labeled data together with a large pool of unlabeled samples. This approach reduces reliance on costly annotations, therefore enabling model to exploit the underlying structure of unlabeled data~\cite{BBAM,PCLC,ssl_1, ssl_2, ssl_3, ssl_4}.\\
\begin{figure}[!t]
\centering
\subfigure[]{\label{fig1a}\includegraphics[width=0.21\textwidth]{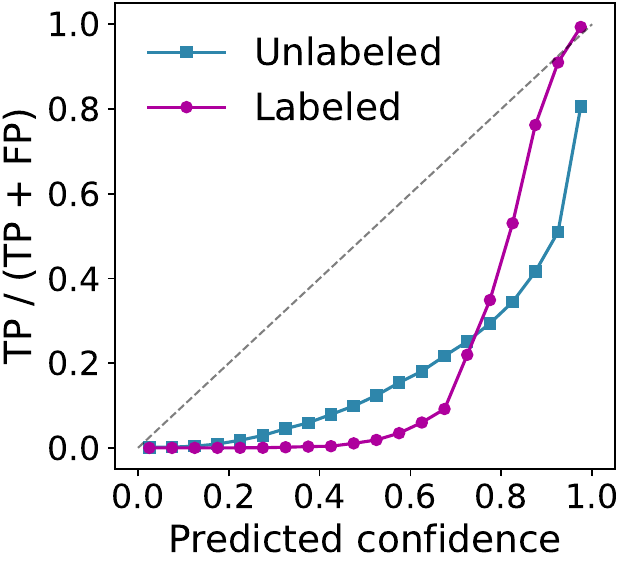}}    
\subfigure[]{\label{fig1b}\includegraphics[width=0.21\textwidth]{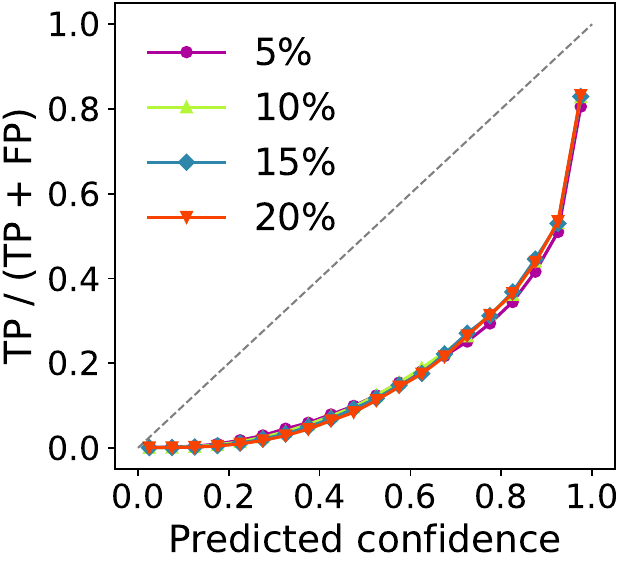}}\\
\subfigure[]{\label{fig1c}\includegraphics[width=0.21\textwidth]{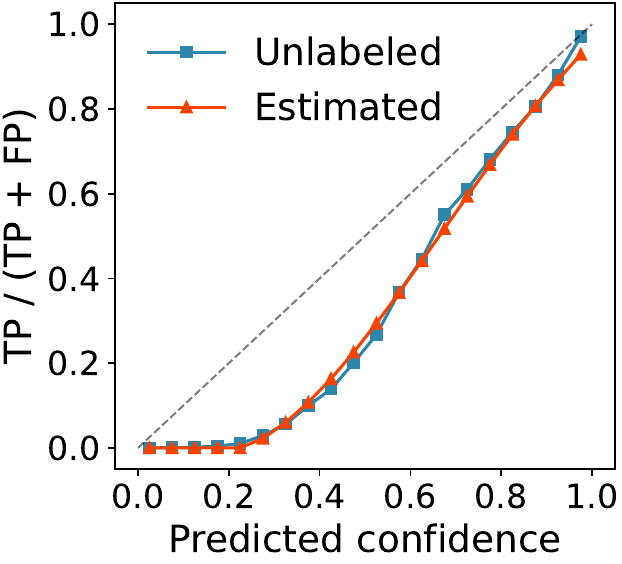}}
\subfigure[]{\label{fig1d}\includegraphics[width=0.21\textwidth]{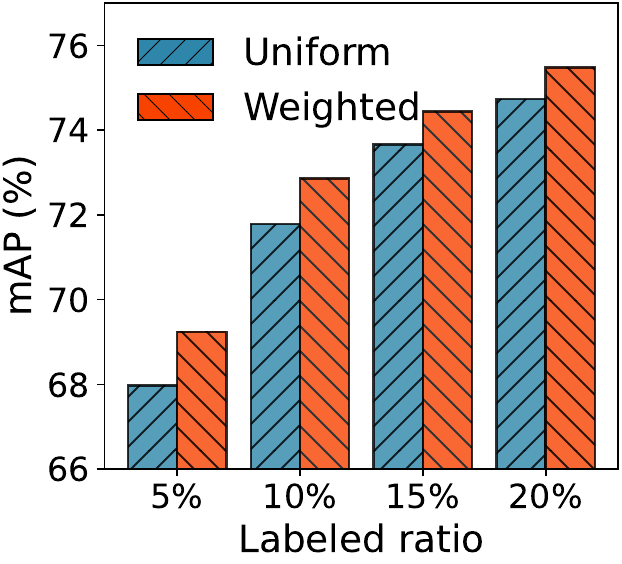}}
\caption{(a) Correctness likelihood distributions of labeled and unlabeled data after warm-up under 5\% labeled setting on COCO. (b) Correctness likelihood distributions of unlabeled data across models trained with varying amounts of labeled data on COCO. (c) Estimated and true distributions for unlabeled data under 5\% labeled setting on VOC. (d) Performance comparison between uniform and correctness-weighted pseudo-labeling on COCO.}
\label{fig1}
\end{figure}
\indent Most existing SSMLL approaches rely on fixed or class-wise thresholds to convert prediction of the unlabeled samples into binary pseudo-labels, and use them to boost the performance. For example, CAP~\cite{CAP} estimates label-wise thresholds by analyzing the positive-to-negative label ratio within the labeled data. Similarly, D2L~\cite{D2L} proposes a metric-adaptive thresholding strategy that considers both the quantity and quality of pseudo-labels. While these methods improve the decision boundaries, they commonly apply \textbf{uniform weights} to all pseudo-labels, regardless of their confidence levels. To be specific, the uniform weighting fails to account for the varying reliability of pseudo-labels, which ultimately limits the performance in semi-supervised multi-label learning. This motivates us to design a more adaptive weighting strategy to better exploit unlabeled data. \\
\indent To this end, we propose a correctness-aware weighting strategy called \underline{\textbf{Di}}stribution-\underline{\textbf{Ca}}librated \underline{\textbf{P}}seudo-labeling~(\textbf{DiCaP}). Our approach is grounded in a theoretically justified insight: the optimal weight assigned to a pseudo-label should reflect its correctness likelihood. Formally, for a pseudo-label identified as positive, its optimal weight should be the proportion of true positives among all predicted positives within the corresponding confidence bin, i.e., $\text{TP} / (\text{TP} + \text{FP})$, where TP and FP denote the number of true and false positive predictions, respectively. Please refer to Section \ref{sec:theoretical-framework} for the detailed theoretical analysis.\\
\indent However, accurately estimating the correctness likelihood is a challenging task. First, as shown in \figurename~\ref{fig1a}, modern deep neural networks tend to produce \textbf{overconfident} predictions, leading to poor calibration between predicted confidence and actual correctness~\cite{calibrate_1}. Second, \figurename~\ref{fig1a} also illustrates \textbf{distinct correctness likelihood distributions} between labeled and unlabeled data due to the difference in supervision signals during training, making it unreliable to estimate the correctness likelihood based on labeled data. \textbf{Fortunately, we observe a consistent phenomenon}: on the same dataset, even when the number of labeled training samples varies, the distribution of pseudo-label correctness likelihood for the unlabeled data remains remarkably stable (see \figurename~\ref{fig1b}). \\ 
\indent This finding motivates a simple yet effective solution: we can randomly split the labeled dataset, e.g., 80\% for supervised training and the remaining 20\% as an estimation set. This estimation subset is first treated as unlabeled and merged with the actual unlabeled data. Since the estimation set and the true unlabeled data are drawn from the same distribution (satisfying the i.i.d. assumption) and are trained under the same strategy, we can reliably estimate the correctness likelihood distribution of pseudo-labels for the unlabeled samples on this set, and dynamically update the weights during training. Despite its simplicity, our estimation strategy demonstrates high accuracy. 
As shown in \figurename~\ref{fig1c}, our estimation method produces a near-perfect match to the true posterior distribution, even under severe label scarcity scenarios (e.g., using only 57 samples for estimation on VOC under 5\% labeled setting). 
The accurate estimation allows the model to assign appropriate weights to pseudo-labels, significantly enhancing model performance, as shown in \figurename~\ref{fig1d}.\\
\indent To further mitigate the impact of ambiguous predictions in pseudo-labeling, we introduce a dual-thresholding strategy that separates class-wise predictions into confident region and highly uncertain (ambiguous) region. For predictions falling within the confident region, we assign pseudo-labels and apply the corresponding estimated weights. In contrast, for predictions that fall into the ambiguous uncertain region, we refrain from applying hard pseudo-labeling and instead apply a consistency-based contrastive loss to regularize their representations in an unsupervised manner. In addition, we repurpose the small estimation subset of labeled data and apply supervised loss during a fine-tuning stage. This final step enables the model to leverage all labeled annotations and preserving accurate pseudo-label calibration throughout the training. \\
\indent Our main contributions are summarized as follows:
\begin{enumerate}
    \item We theoretically establish that, the optimal weight assigned to any pseudo-label equals to its correctness likelihood in SSMLL. 
    \item We observe that on the same dataset, the correctness likelihood distribution of unlabeled samples remains consistent, even under different numbers of labeled training samples. Leveraging this insight, a simple data-splitting approach was proposed to accurately estimate the weight distribution of the unlabeled samples.
    \item We propose a dual-thresholding strategy that separates predictions into confident and uncertain regions, and adopts different strategies to handle each region.
    \item Extensive experimental results validate that our method outperforms the current state-of-the-art~(SOTA) methods by a large margin, e.g., a \textbf{4.27\%} improvement on VOC with limited labeled data.
\end{enumerate}
\section{Related Work}
\subsection{Multi-Label Learning}
\indent  Multi-label learning addresses the problem where each instance can be associated with multiple labels. Modeling label correlations has become a mainstream research direction in multi-label learning. To effectively model label dependencies, neural networks with various architectures have been proposed from various perspectives, including recurrent neural networks~\cite{cnn-rnn}, graph convolutional networks~\cite{ML-GCN}, and Transformer architectures~\cite{Query2Label}. On the other hand, many methods~\cite{ASL, APP_longtail} focus on addressing label imbalance issues in multi-label learning, including both intra-class imbalance (i.e., imbalance between positive and negative instances within each class) and inter-class imbalance (i.e., long-tailed label distributions). For instance, Ridnik et al. proposed Asymmetric Loss (ASL), which dynamically down-weights easy negatives samples and introduces hard thresholding~\cite{ASL}.\\
 \indent Given the high cost and effort required to obtain fully annotated labels for multi-label data, recent research has increasingly focused on weakly supervised variants of MLL. These include semi-supervised multi-label learning~\cite{D2L}, multi-label learning with missing labels~\cite{ref_cam}, and partial multi-label learning~\cite{ref_pml_cd, PML_fuchao}, all aiming to reduce annotation burden while maintaining competitive performance.
 \subsection{Semi-Supervised Multi-Label Learning}
\indent SSMLL tackles the challenge of learning with limited labeled data by leveraging additional unlabeled samples. Various methods have been proposed to generate reliable pseudo-labels or effectively utilize unlabeled data in SSMLL. For instance, Xie et al. adopted a class-specific thresholding approach, where thresholds are dynamically estimated from class priors to binarize predictions for unlabeled instances~\cite{CAP}. Liu et al. enhanced pseudo-label quality by introducing a causality-guided label prior, inferred via a Structural Causal Model, and incorporates this prior into a variational inference framework to guide label enhancement~\cite{PCLC}. Li et al. addressed the variance bias between positive and negative samples by extending the binary angular margin loss with Gaussian-based feature angle transformations, and introduces a prototype-aware negative sampling scheme to further stabilize training~\cite{BBAM}. Xiao et al. adopted a dual-decoupling strategy to jointly model discriminative and correlative information, and introduces a metric-adaptive thresholding mechanism to dynamically refine pseudo-label assignment throughout training~\cite{D2L}.\\
\indent However, these methods assign \textbf{uniform weights} to all pseudo-labels without considering their quality, causing low-quality pseudo-labels to harm performance. Therefore, studying how to obtain better pseudo-label weights is important. In the following section, we theoretically analyze the optimal weighting scheme.
\section{Theoretical Analysis}
\label{sec:theoretical-framework}
\indent To further justify the design of our correctness-aware weighting scheme, we theoretically characterize the impact of pseudo-label noise on the learning objective in SSMLL.
\subsection{Problem Formulation}
\label{sec:preliminaries_and_formulation}
\indent We begin by formally defining the problem setting and notation for SSMLL. Let $x \in \mathcal{X}$ denote a feature vector and $y \in \mathcal{Y}$ denote the corresponding multi-label annotation, where $\mathcal{X} \subseteq \mathbb{R}^d$ is the input space and $\mathcal{Y} = \{0, 1\}^C$ is the label space consisting of $C$ possible categories. And, $y_{ic} = 1$ indicates the relevance of the $c$-th label to the instance $x_i$, while $y_c = 0$ signifies irrelevance.\\
\indent In SSMLL, the training data consists of a labeled set with $N_l$ training examples $\mathcal{D}_l = \{(x_i, y_i)\}_{i=1}^{N_l}$ and an unlabeled set with $N_u$ training instances $\mathcal{D}_u = \{x_j\}_{j=1}^{N_u}$, where $N_l \ll N_u$. The primary objective is to train a multi-label classifier by leveraging both labeled and unlabeled data, where the unlabeled data are assigned pseudo-labels $\hat y$. The model, parameterized by $\theta$ and denoted as $f_{\theta}: \mathcal{X} \rightarrow [0,1]^C$, predicts a vector of class-wise relevance scores. 
\subsection{Derivation of the Optimal Weighting Strategy}
\indent In SSMLL, model training heavily depends on the pseudo-labels $\hat{y}$ assigned to the unlabeled data. However, these pseudo-labels are inevitably noisy. An effective strategy is to assign a weight to the loss of each pseudo-labeled sample, so that more reliable predictions are emphasized and less reliable ones are down-weighted. Intuitively, this weight $w(p)$ should be a function of the model's confidence $p$.\\
\indent To derive the optimal weighting function, we assume the ground-truth labels of the unlabeled data are known. Let $\mathbb{I}[\hat{y} = y]$ be a binary indicator denoting whether a pseudo-label is correct. The goal is to assign higher weights to correct pseudo-labels and lower weights to incorrect ones. This leads to formulating the optimization of $w(p)$ as minimizing the binary cross-entropy (BCE) loss between the weight and the correctness indicator. For a pseudo-label $\hat{y}_i$ with confidence $p_i$ and ground-truth label $y_i$, the loss is defined as:
  \begin{align}
    \ell_\text{BCE}\left(w(p_i), \mathbb{I}[\hat y_i = y_i] \right) 
     = & \underbrace{- \mathbb{I}[\hat y_i = y_i] \log ( w(p_i))}_\text{increase weight for correct samples} \label{eq:bce:sample}\\
     & \underbrace{- \mathbb{I}[\hat y_i \neq y_i] \log (1 - w(p_i))}_\text{decrease weight for incorrect samples}. \nonumber
  \end{align}
\indent Since a single confidence value $p_i$ typically corresponds to multiple samples, we aim to minimize the expected BCE loss over all samples with same confidence:
  \begin{equation}
  \label{eq:optimal-weight}
  \begin{aligned}
    w^{*}(p_i) = & \arg \min_{w(p_i)} \ \mathbb{E}_{y \mid p_i} \left[ \ell_\text{BCE}(w(p_i),\mathbb{I}[\hat y =  y]) \right] \\
    =  & P(\hat y =  y \mid p_i),
  \end{aligned}
  \end{equation}
which shows that the optimal weighting function is exactly the posterior correctness likelihood of the pseudo-label given the confidence score $p_i$.\\
\indent However, estimating correctness likelihood at each individual confidence value $p_i$ is inherently unreliable due to data sparsity—especially under limited data, where only a few samples may share the same confidence score. To address this issue, we propose to approximate the optimal weighting function by aggregating samples within similar confidence ranges. The continuous confidence interval $[0, 1]$ is divided into $K$ non-overlapping bins, denoted as $\{\mathcal{B}_1, \dots, \mathcal{B}_K\}$. For each bin $\mathcal{B}_k$, we compute the empirical correctness rate of pseudo-labels whose predicted confidence scores fall within that bin. The optimal correctness likelihood on unlabeled data is then given by:
\begin{equation}
  w^{*}_k(p) = P(\hat{y} = y \mid p \in \mathcal{B}_k),
\end{equation}
which serves as a practical approximation of the optimal weight for any confidence value $p \in \mathcal{B}_k$.
\section{Methods}
\begin{figure}
    \centering
    \includegraphics[width=0.85\linewidth]{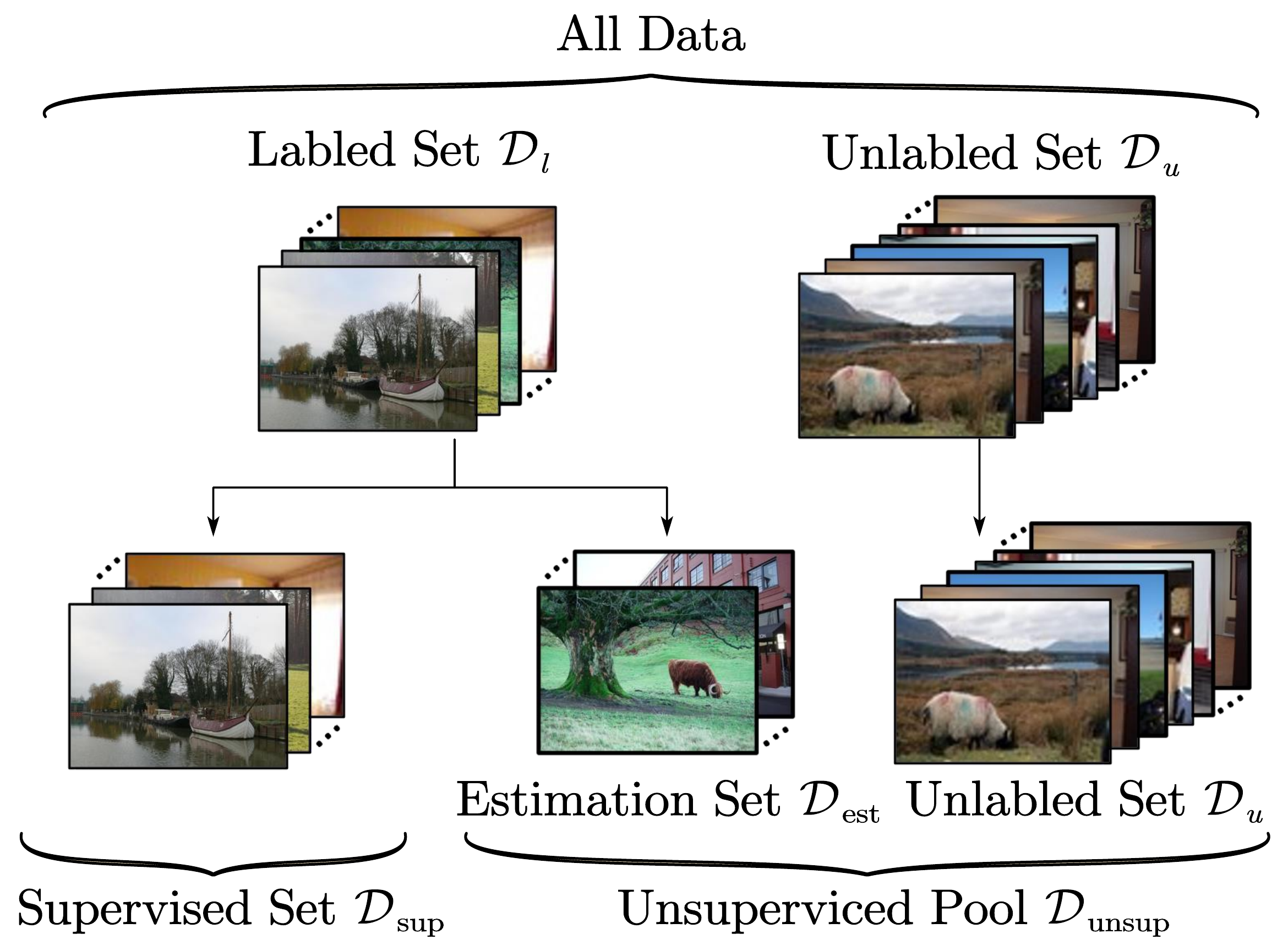}
    \caption{Illustration of estimation set construction and relationships among $\mathcal{D}_l$, $\mathcal{D}_u$, $\mathcal{D}_{\text{sup}}$, $\mathcal{D}_{\text{est}}$, and $\mathcal{D}_{\text{unsup}}$.}
    \label{fig:esc}
\end{figure}

\subsection{Distribution-Calibrated Weight Estimation}
\subsubsection{Estimation Set Construction.}
\indent As discussed earlier, the optimal weight for a pseudo-label should reflect its correctness likelihood. However, estimating this likelihood is inherently challenging without ground-truth labels for unlabeled data. Moreover, due to overconfidence issue and the different correctness distributions between labeled and unlabeled data, it is unreliable to infer correctness likelihood directly from the model's predicted probabilities or based on the labeled set. Fortunately, we observe that within the same dataset, the distribution of correctness likehood over unlabeled data remains stable, even when the number of labeled training samples varies (see Figure~\ref{fig1b}).\\ 
\indent Motivated by this observation, we divide the labeled dataset $\mathcal{D}_l$ into two disjoint subsets: a supervised training set $\mathcal{D}_{\text{sup}}$ for supervised training and an estimation set $\mathcal{D}_{\text{est}}$ for estimating the weight, such that $\mathcal{D}_l = \mathcal{D}_{\text{sup}} \cup \mathcal{D}_{\text{est}}$ and $\mathcal{D}_{\text{sup}} \cap \mathcal{D}_{\text{est}} = \emptyset$. The estimation set $\mathcal{D}_{\text{est}}$ is treated as unlabeled and merged with the original unlabeled data $\mathcal{D}_u$ to form a unified pool:
\begin{equation}
\mathcal{D}_{\text{unsup}} = \mathcal{D}_u \cup \{x_i \mid (x_i, y_i) \in \mathcal{D}_{\text{est}}\}.
\end{equation}
\indent As shown in \figurename~\ref{fig:esc}, samples from $\mathcal{D}_{\text{est}}$ are handled identically to those in $\mathcal{D}_u$, ensuring distributional consistency across the entire unsupervised pool. This design allows us to reliably use $\mathcal{D}_{\text{est}}$ to estimate the correctness likelihood distribution over the unlabeled data.
\subsubsection{Weight Estimation.}
\indent As discussed above, we can empirically approximate the weight using the estimation set $\mathcal{D}_{\text{est}}$. Here, we introduce the specific implementation strategy for distribution-calibrated weight estimation for $\mathcal{D}_{\text{unsup}}$.\\
\indent Following the theoretical analysis in Section \ref{sec:theoretical-framework}, we uniformly partition the confidence interval $[0, 1]$ into $K=20$ non-overlapping bins, denoted as $\{\mathcal{B}_1, \dots, \mathcal{B}_K\}$, where each bin corresponds to the range $\mathcal{B}_k = \left[\frac{k-1}{K}, \frac{k}{K}\right)$. For each prediction score $p$ and its associated ground-truth label $y$ in $\mathcal{D}_{\text{est}}$, we count the number of true positives and true negatives that fall into each confidence bin:
\begin{equation}
\begin{aligned}
n_k^{\text{pos}} &= \sum_{(x, y) \in \mathcal{D}_{\text{est}}} \sum_{c=1}^{C} \mathbb{I}\left(p_{c} \in \mathcal{B}_k,\; y_{c} = 1\right), \\
n_k^{\text{neg}} &= \sum_{(x, y) \in \mathcal{D}_{\text{est}}} \sum_{c=1}^{C} \mathbb{I}\left(p_{c} \in \mathcal{B}_k,\; y_{c} = 0\right).
\end{aligned}
\end{equation}
\indent We then compute the estimated positive and negative proportions as:
\begin{equation}
r_k^{\text{pos}} = \frac{n_k^{\text{pos}}}{n_k^{\text{pos}} + n_k^{\text{neg}} + \epsilon}, \qquad r_k^{\text{neg}} = 1 - r_k^{\text{pos}},
\end{equation}
where \(\epsilon\) is a small constant to avoid division by zero.\\
\indent During training, for any predicted confidence \(p\) and corresponding pseudo-label \(\hat{y} \in \{0,1\}\), we assign a soft weight based on linear interpolation between adjacent bins:
\begin{align}
w(p , \hat{y} = 1) &= \left(\frac{k+1}{K} - p\right) r_k^{\text{pos}} + \left(p - \frac{k}{K}\right) r_{k+1}^{\text{pos}}, \nonumber\\
w(p , \hat{y} = 0) &= \left(\frac{k+1}{K} - p\right) r_k^{\text{neg}} + \left(p - \frac{k}{K}\right) r_{k+1}^{\text{neg}}, \nonumber\\
\text{s.t.} \quad & \frac{k}{K} \leq p < \frac{k+1}{K}.\label{eq:weight_compute}
\end{align}
\indent This correctness-based weight reflects the estimated probability that the pseudo-label is correct, and is directly used to modulate the training loss.

\subsection{Dual-Thresholding for Pseudo-Label}

While correctness-based weighting alleviates the effect of noisy pseudo-labels, there still exist highly ambiguous regions where predictions are particularly unreliable, which can negatively affect model training. To further mitigate the impact of such uncertain predictions, we adopt a dual-thresholding strategy to explicitly partition pseudo-labeled data into confident and uncertain subsets.

Inspired by~\cite{DESP}, we derive class-wise dynamic thresholds from the prediction statistics on the labeled training set. For each class \(c\), we first collect the predicted confidence scores from positive and negative label in \(\mathcal{D}_{\text{sup}}\). Let \(\mathcal{P}_c^{\text{pos}} = \{ p_{ic} \mid (x_i, y_i) \in \mathcal{D}_{\text{sup}},\; y_{ic}=1 \}\) and \(\mathcal{P}_c^{\text{neg}} = \{ p_{ic} \mid (x_i, y_i) \in \mathcal{D}_{\text{sup}},\; y_{ic}=0 \}\). We define the thresholds as the mid-range of each group:
\begin{equation}
\begin{aligned}
\tau_c^{\text{pos}} &= \frac{\max(\mathcal{P}_c^{\text{pos}}) + \min(\mathcal{P}_c^{\text{pos}})}{2}, \\
\tau_c^{\text{neg}} &= \frac{\max(\mathcal{P}_c^{\text{neg}}) + \min(\mathcal{P}_c^{\text{neg}})}{2}.
\end{aligned}
\end{equation}

Given an unlabeled sample \(x_u\) and its predicted confidence scores \(\{p_{uc}\}_{c=1}^{C}\), we assign pseudo-labels by applying the derived dual thresholds for each class. The pseudo-label \(\hat{y}_{uc}\) is categorized into one of three types based on its confidence score:
\begin{equation}
    \hat{y}_{uc} = \begin{cases}
    1, \quad &p_{uc} > \tau_c^{\text{pos}}, \\
    0, \quad &p_{uc} < \tau_c^{\text{neg}},\\
    -1, \quad &\tau_c^{\text{neg}} \leq p_{uc} \leq \tau_c^{\text{pos}},
    \end{cases}    
\label{pseudo-label}
\end{equation}
where the labels \(1\), \(0\), and \(-1\) correspond to confident positive, confident negative, and uncertain predictions, respectively. The samples assigned with \(1\) or \(0\) are aggregated into the confident set $\mathcal{D}_{\text{conf}}$, while those assigned with \(-1\) constitute the uncertainty set $\mathcal{D}_{\text{uncer}}$.

We retain only the confident samples $(x_{ic},\hat{y}_{ic}) \in \mathcal{D}_{\text{conf}}$ for supervised training. Their pseudo-label loss is computed as:
\begin{equation}
\mathcal{L}_{\text{pseudo}} = \frac{1}{|\mathcal{D}_{\text{conf}}|} \sum_{\hat{y}_{ic} \in \{1,0\}}
w(p_{ic}, \hat{y}_{ic}) \cdot \ell(p_{ic}, \hat{y}_{ic}),
\end{equation}
where \(w(\cdot)\) denotes the correctness-based weight defined in Eq. (\ref{eq:weight_compute}), and we use Asymmetric Loss~(ASL)~\cite{ASL} for $\ell(\cdot,\cdot)$. 

\subsection{Robust Representation Learning for Uncertain Samples}
\indent Uncertain samples, whose predictions lie within the ambiguous confidence interval, are unsuitable for direct supervision. To exploit their representational value without introducing noisy gradients, we incorporate an unsupervised contrastive learning objective tailored to the multi-label setting. \\
\indent Specifically, we extend the standard instance-level InfoNCE loss~\cite{InforNCE} into a \textbf{class-wise contrastive framework} to suit the multi-label setting. For each sample \(x_i\), we generate a weakly augmented view \(x_i^w\) and a strongly augmented view \(x_i^s\). A shared encoder is used to extract class-wise feature embeddings \(\{z_{ic}^w\}_{c=1}^{C}\) and \(\{z_{ic}^s\}_{c=1}^{C}\) for the two views, respectively. For all uncertain predictions (i.e., those with \(\hat{y}_{ic} = -1\)), we collect the corresponding feature pairs \((z_{ic}^w, z_{ic}^s)\) to construct the contrastive training set.\\
\indent We treat each pair \((z_{ic}^w, z_{ic}^s)\) as a positive pair and all other class-wise features within the mini-batch as negatives. The class-wise contrastive loss is defined as:
\begin{equation}
\mathcal{L}_{\text{uncer}} = -\frac{1}{2B} \sum_{i=1}^{2B} \log \frac{\exp(z_i \cdot z_i^+ / \tau)}{\sum_{j=1}^{2B} \mathbb{I}_{i \neq j}\, \exp(z_i \cdot z_j / \tau)},
\end{equation}
where \(B\) denotes the number of uncertain samples, \(z_i^+\) is the positive counterpart of \(z_i\), and \(\tau\) is a temperature hyperparameter that controls the sharpness of the distribution. Inspired by recent self-supervised learning study~\cite{self_learning}, we can also employ it to utilize all data during the warm-up stage to improve representation generality.\\
\indent Finally, the overall training objective during the pseudo-labeling stage is defined as:
\begin{equation}
\mathcal{L} = \mathcal{L}_{\text{sup}} + \mathcal{L}_{\text{pseudo}} + \mathcal{L}_{\text{uncer}},
\end{equation}
where $\mathcal{L}_{\text{sup}}$ denotes the supervised loss on $\mathcal{D}_{\text{sup}}$, $\mathcal{L}_{\text{pseudo}}$ is the weighted pseudo-label loss on confident samples, and $\mathcal{L}_{\text{uncer}}$ represents the contrastive loss of uncertain samples.

\subsection{Fine-Tuning on Estimation Set}

To fully utilize all labeled data, we introduce a final fine-tuning stage on the estimation subset $\mathcal{D}_{\text{est}}$. This subset was previously treated as unlabeled to estimate pseudo-label reliability, and its ground-truth annotations were excluded from earlier supervised training. Leveraging its ground-truth annotations can thus further enhance the model's generalization on all labeled data.

To mitigate overfitting and reduce computational cost, we freeze the backbone and fine-tune only the classification head using $\mathcal{D}_{\text{est}}$. The optimization objective is:
\begin{equation}
\mathcal{L}_{\text{ft}} = \frac{1}{|\mathcal{D}_{\text{est}}|} \sum_{(x, y) \in \mathcal{D}_{\text{est}}} \ell(f_\theta(x), y),
\label{total_loss}
\end{equation}
where $\ell(\cdot,\cdot)$ denotes ASL. By incorporating previously unused annotations, this fine-tuning stage complements the training process and ensures complete utilization of labeled data.

For ease of understanding, we provide the algorithmic process in Algorithm \ref{alg:algorithm}.

\begin{algorithm}
  \caption{Algorithm of DiCaP}
  \label{alg:algorithm}
  \textbf{Input}: Labeled dataset $\mathcal{D}_l$, Unlabeled dataset $\mathcal{D}_u$ \\
 \textbf{Output}: Trained multi-label classifier $f_\theta$
  
  \begin{algorithmic}[1]
    \STATE Split $\mathcal{D}_l$ into $\mathcal{D}_{\text{sup}}$ and $\mathcal{D}_{\text{est}}$, $\mathcal{D}_{\text{unsup}} = \mathcal{D}_u \cup \mathcal{D}_{\text{est}}$.
    \STATE Warm up on $\mathcal{D}_{\text{sup}} \cup \mathcal{D}_{\text{unsup}}$ to initialize $f_\theta$.
    \FOR{each epoch}
      \STATE Estimate pseudo-label weights $w(p)$ using $\mathcal{D}_{\text{est}}$.
      \STATE Derive thresholds $\{\tau_c^{\text{pos}},\tau_c^{\text{neg}}\}^C$ from $\mathcal{D}_{\text{sup}}$.
      \STATE Generate pseudo-labels for $\mathcal{D}_{\text{unsup}}$ using Eq.~(\ref{pseudo-label}).
      \STATE Update model by minimizing the total loss Eq.~(\ref{total_loss}).
    \ENDFOR
    \STATE Fine-tune the classification head on $\mathcal{D}_{\text{est}}$.
  \end{algorithmic}
\end{algorithm}

  \begin{table*}[tb]
    \renewcommand{\arraystretch}{1.0}
    \small
    \setlength\tabcolsep{2.8pt}
    \centering
    \begin{tabular}{lcccccccccccccccc}
    \toprule
     \multirow{2}{*}{Method} & \multicolumn{4}{c}{VOC} & \multicolumn{4}{c}{COCO} & \multicolumn{4}{c}{NUS} & \multicolumn{4}{c}{AWA} \\
      \cmidrule(r){2-5}\cmidrule(r){6-9}\cmidrule(r){10-13}\cmidrule(r){14-17}
       & $5\%$& $10\%$& $15\%$& $20\%$& $5\%$& $10\%$& $15\%$& $20\%$& $5\%$& $10\%$& $15\%$& $20\%$& $5\%$& $10\%$& $15\%$& $20\%$\\
      \midrule
      BCE        & $65.40$&$75.48$&$77.87$&$79.00$& $57.09$&$62.34$&$65.55$&$67.31$& $40.12$&$45.04$&$47.04$&$48.29$& $61.33$&$63.35$&$62.51$&$63.00$\\
      ASL        & $71.41$&$77.81$&$79.12$&$79.84$& $57.87$&$62.95$&$65.73$&$67.43$& $42.04$&$46.07$&$48.04$&$49.55$& $60.40$&$63.06$&$62.88$&$63.31$\\
      MLD  & $74.18$&$81.10$&$82.74$&$84.29$& $62.45$&$67.27$&$69.95$&$71.22$& $39.78$&$44.31$&$47.02$&$49.06$& $62.28$&$63.57$&$63.17$&$63.94$\\
\midrule
 Top-1& $75.34$& $80.80$& $82.93$& $83.67$& $59.42$& $63.52$& $65.13$& $66.88$& $39.27$& $46.05$& $47.02$& $46.69$& $63.39$& $64.63$& $64.60$&$63.86$\\
 Top-k& $73.62$& $80.20$& $82.17$& $83.03$& $59.83$& $64.02$& $65.21$& $67.45$& $39.18$& $46.15$& $46.98$& $46.70$& $63.87$& $64.41$& $64.81$&$64.49$\\
 IAT&   $71.88$& $80.18$& $82.87$& $83.99$& $60.76$& $65.60$& $65.31$& $69.29$& $40.10$& $46.45$& $47.39$& $47.15$& $62.93$& $64.17$& $63.57$&$63.57$\\ \midrule
    DRML & $61.75$&$70.97$&$72.97$&$74.44$& $53.59$&$57.02$&$58.62$&$59.18$& $30.57$&$35.03$&$37.93$&$40.01$& $61.60$&$62.46$&$63.77$&$63.38$\\
    CAP  & $75.90$& $81.83$& $83.10$& $84.32$& $62.88$& $67.18$& $68.99$& $70.43$& $44.98$& $47.81$& $49.04$& $51.37$& $63.90$& $64.15$& $64.40$&$64.51$\\
    PCLP & $77.25$& $82.21$& $83.72$& $84.59$& $64.43$& $69.02$& $70.86$& $71.52$& $46.39$& $48.83$& $50.57$& $52.45$& $64.30$& $65.38$& $64.47$&$64.49$\\
    BBAM & $78.66$&$83.45$&$84.54$&$84.58$& $63.54$&$67.41$&$68.86$&$70.23$& $33.15$&$39.12$&$41.26$&$42.52$& $64.19$&$64.63$&$64.16$&$64.76$\\
    D2L  & $\underline{79.26}$&$\underline{84.06}$&$\underline{86.25}$&$\underline{87.16}$& $\underline{69.30}$&$\underline{73.06}$&$\underline{74.63}$&$\underline{75.70}$&$\underline{46.86}$&$\underline{50.25}$&$\underline{51.61}$&$\underline{52.64}$& $\underline{64.66}$&$\underline{65.57}$&$\underline{64.95}$&$\underline{65.06}$\\
    \midrule
      \rowcolor{gray!30} Ours&$\mathbf{83.53}$ &$\mathbf{87.92}$&$\mathbf{88.40}$&$\mathbf{88.48}$& $\mathbf{70.07}$&$\mathbf{73.55}$&$\mathbf{75.06}$&$\mathbf{75.90}$& $\mathbf{48.37}$&$\mathbf{51.24}$&$\mathbf{52.30}$&$\mathbf{53.61}$& $\mathbf{66.32}$&$\mathbf{66.48}$&$\mathbf{66.49}$&$\mathbf{66.78}$
\\
\rowcolor{gray!30} $\Delta$ &$+4.27$ & $+3.86$& $+2.15$& $+1.32$& $+0.77$& $+0.49$& $+0.43$& $+0.20$& $+1.51$& $+0.99$& $+0.69$& $+0.97$& $+1.66$& $+0.91$& $+1.54$&$+1.72$
\\ \bottomrule
    \end{tabular}%
    \caption{Comparison of our model with other methods, with labeled rate $5\%$, $10\%$, $15\%$ and $20\%$. mAP($\%$) is adopted as the evaluation metric. Optimal values are denoted in bold, and the second-best values are underlined. $\Delta$ denotes the performance gap between our method and the best compared method.}
    \label{tab:main-result}%
  \end{table*}

\section{Experiments}

\subsection{Settings}
 \subsubsection{Datasets.} 
 \indent To validate our proposed approach, we conducted experiments on four commonly used real-world multi-label image datasets, including MS-COCO 2014 (COCO) \cite{COCO}, Pascal VOC 2007 (VOC) \cite{VOC}, NUS-WIDE (NUS) \cite{NUS-WIDE}, and Animals with Attributes2 (AWA) \cite{AWA}. Following \cite{CAP}, we transform these datasets into SSL versions. For each dataset, we randomly select a proportion $\rho$ of training samples as labeled ones, and the remaining as unlabeled ones. We set $\rho \in \{5\%, 10\%, 15\%, 20\%\}$, to explore the performance of our method under different data proportions.   
 \subsubsection{Comparison Methods.} 
\indent We compare our model with 11 baseline methods, which are categorized into three groups. 
\textbf{BCE}, \textbf{ASL}~\cite{ASL}, and \textbf{MLD}~\cite{MLdecoder} are classic multi-label learning methods. These methods only use the labeled data for training.
\textbf{Top-1}, \textbf{Top-k}, and \textbf{IAT}~\cite{CAP} are instance-based pseudo-labeling methods.
\textbf{DRML}~\cite{DRML}, \textbf{CAP}~\cite{CAP}, \textbf{PCLP}~\cite{PCLC}, \textbf{BBAM}~\cite{BBAM}, and \textbf{D2L}~\cite{D2L} represent the SOTA SSMLL methods based on deep models.
\subsubsection{Implementation Details.} 
\indent Following the experimental setup in \cite{CAP}, we adopt a ResNet-50 \cite{ResNet} backbone pre-trained on ImageNet \cite{imagenet} for feature extraction. For the decoder, we employ the ML-Decoder framework \cite{MLdecoder} to produce class-wise embeddings. All input images are resized to $224 \times 224$, and data augmentation is applied using RandAugment \cite{Randaugment} and Cutout \cite{cutout}. The optimizer is AdamW \cite{adamw} with the weight decay of $1e-4$, and we employ a one-cycle learning rate scheduler with a maximum learning rate of $1e-4$. To stabilize training, we apply an exponential moving average (EMA) to the model parameters with a decay rate of 0.9997. All experiments are conducted on NVIDIA GeForce RTX 3090 GPUs. To ensure reproducibility, the random seed is fixed to 1 throughout all experiments. Consistent with prior works~\cite{CAP}, mean average precision (mAP) is applied to evaluate models’ performance. \\
\indent For all compared methods, we use their officially recommended optimal hyperparameter settings. During the warm-up phase of our method, we utilize 80\% of the labeled data for supervised learning. In contrast, all baseline methods follow their original protocols and leverage the full set of labeled data during training. In the fine-tuning phase, the backbone is frozen, and only the classification head is updated. We train for 20 epochs with a batch size of 32. 
\subsection{Comparison with SOTA Methods}
\indent As shown in Table~\ref{tab:main-result}, we conduct a comprehensive comparison against SOTA methods across four benchmark datasets under four different labeled data ratios. The results clearly demonstrate that our proposed method consistently achieves the best performance across all datasets and label-sparsity settings. Specifically, our method surpasses the second-best approach by an average margin of 2.9\% on the VOC dataset. Our method also remains robust in highly challenging scenarios where only 5\% of the training data is labeled. Under this limited supervision setting, we outperform the second-best method by notable margins with 4.27\%, 1.51\%, and 1.67\% on VOC, NUS-WIDE, and AWA, respectively, highlighting the effectiveness of our method.\\
\indent Among the baselines, DRML, a two-stage semi-supervised method, performs poorly on large-scale image datasets. This is mainly due to its decoupled training scheme, which prevents end-to-end feature learning. As a result, DRML performs worse than baselines BCE and ASL. On the other hand, D2L achieves consistently competitive results across all datasets, securing the second-best performance overall. This demonstrates the benefit of refining pseudo-labels by leveraging model predictions. However, D2L applies the uniform weight to all pseudo-labels, which limits its performance. By comparison, our method explicitly models the \textit{correctness likelihood} of each pseudo-label and dynamically adjusts their contributions during training. This principled weighting mechanism enables a more effective use of pseudo-labels and leads to consistent performance improvements.
\begin{figure*}
\centering
\subfigure[]{  \includegraphics[width=0.157\textwidth]{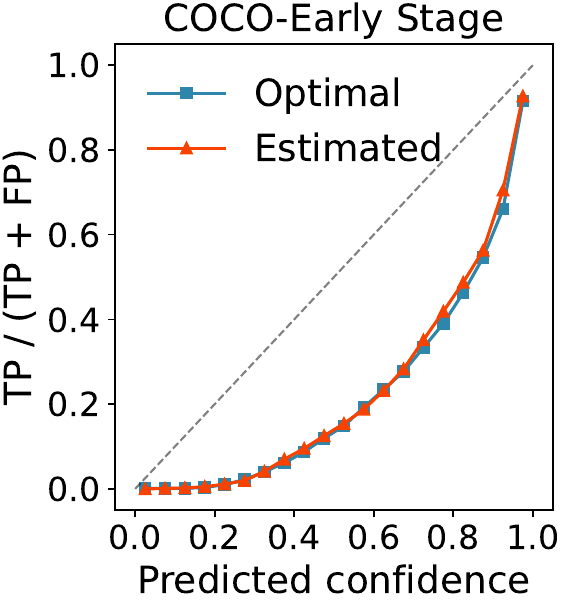}}
\subfigure[]{  \includegraphics[width=0.157\textwidth]{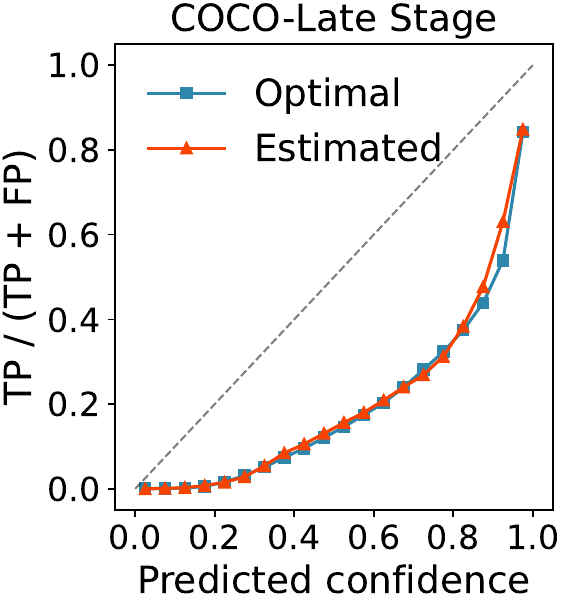}}
\subfigure[]{  \includegraphics[width=0.157\textwidth]{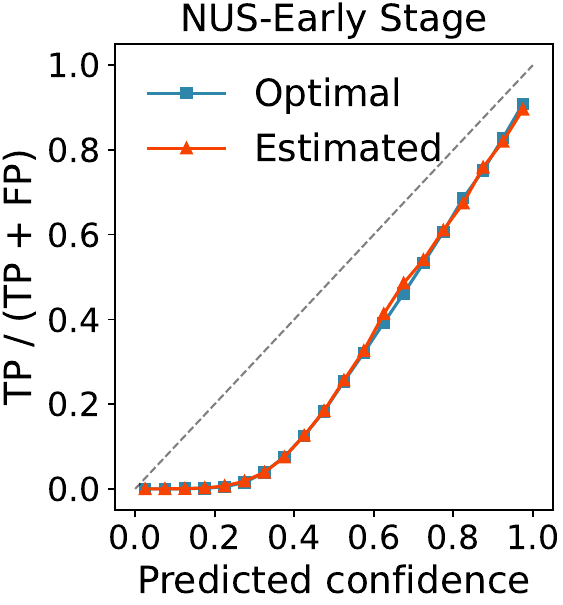}}
\subfigure[]{  \includegraphics[width=0.157\textwidth]{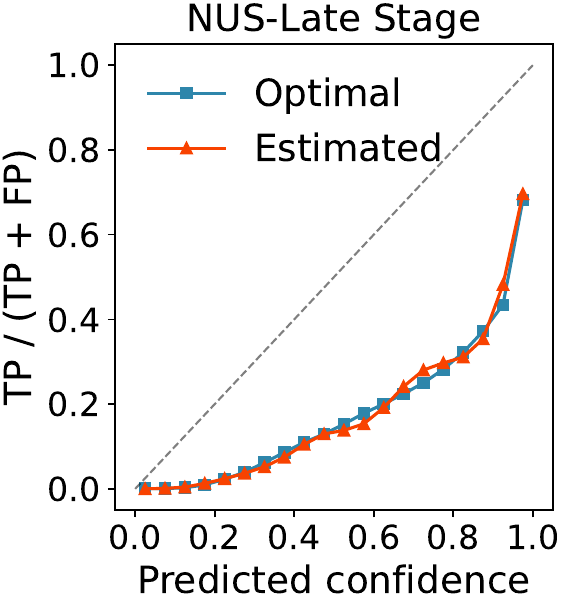}}
\subfigure[]{  \includegraphics[width=0.157\textwidth]{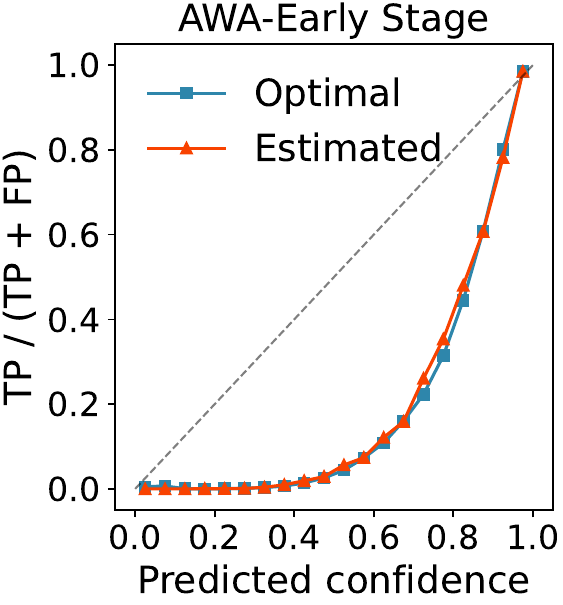}}
\subfigure[]{  \includegraphics[width=0.157\textwidth]{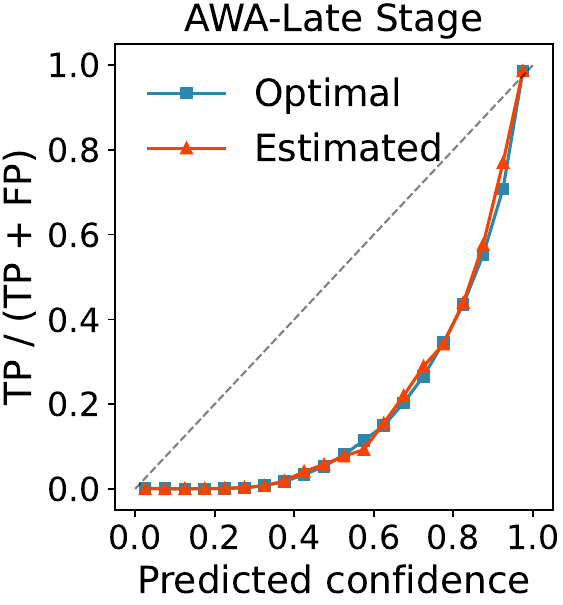}}
\caption{Comparison between estimated and optimal correctness likelihood distributions at early and late training stages on COCO, NUS and AWA under 5\% labeled setting.}
  \label{fig:test_weight}
\end{figure*}

\subsection{Further Analysis}
 \begin{table}[t]
    \footnotesize
    \setlength\tabcolsep{5pt}
    \centering
    \begin{tabular}{lccccc}
    \toprule
    \multirow{2}{*}{Ablations }  & \multicolumn{2}{c}{COCO} & \multicolumn{2}{c}{NUS} & \multirow{2}{*}{Avg.}\\
      \cmidrule(r){2-3} \cmidrule(r){4-5}
     & $5\%$ & $10\%$ & $5\%$ & $10\%$   & \\
    \midrule      
       Baseline       & $65.13$&$69.53$&$42.69$&$45.91$ & $55.82$$~~~~~~~~~$\\
    \midrule      
        + DCW   & $69.18$&$72.46$&$45.20$&$48.08$ & $58.73$$ \scriptscriptstyle ~\mathbf{+2.91}$\\
    + DTH   & $69.24$& $72.86$& $47.73$&$49.82$ & $59.91$$ \scriptscriptstyle ~+1.18$\\
    + URRL& $69.79$&$72.97$&$47.86$&$50.38$ & $60.25$$\scriptscriptstyle ~+0.34$\\    
    + WCL& $69.98$&$73.14$& $47.96$&$50.51$ & $60.40$$\scriptscriptstyle ~+0.15$\\
    + FTE   & $\mathbf{70.07}$& $\mathbf{73.55}$& $\mathbf{48.37}$&$\mathbf{51.24}$& $\mathbf{60.81}\scriptscriptstyle ~+0.41$
\\
  \bottomrule
    \end{tabular}%
    \caption{Ablation study of different components on COCO and NUS. \textbf{Baseline} corresponds to training only on labeled data. \textbf{DCW} denotes using the estimated correctness weights as soft pseudo-labels. \textbf{DTH} refers to the dual-threshold confidence partitioning strategy. \textbf{URRL} represents robust representation learning for uncertain samples. \textbf{WCL} indicates incorporating all data into the warm-up stage via class-wise contrastive loss. \textbf{FTE} denotes fine-tuning on $\mathcal{D}_{\text{est}}$. mAP(\%) is adopted as metric.}
    \label{tab:ablation-study}%
  \end{table}

\subsubsection{All components contribute, with the distribution-calibrated weighting providing the largest gain.}
\indent In Table~\ref{tab:ablation-study}, we evaluate the contribution of each component in DiCaP through ablation experiments on COCO and NUS under 5\% and 10\% labeled data. The results clearly demonstrate that all modules contribute positively to the overall performance. For instance, on COCO with 5\% labeled data, incorporating the distribution-calibrated weighting~(DCW) module brings a notable improvement of 4.05\%. On NUS, introducing the dual-thresholding strategy~(DTH) yields an additional 2.14\% gain. The benefit of the final fine-tuning step becomes more evident as the proportion of labeled data increases, providing a 0.73\% improvement on NUS with 10\% labeled data.
Among all components, the DCW module delivers the most significant overall contribution, with an average gain of 2.92\%. These consistent improvements highlight the importance of distribution-calibrated weighting strategy.
\subsubsection{Our estimated correctness likelihoods are highly accurate.}
\indent To further validate the accuracy of our estimated correctness-likelihood distributions, we visualize both the estimated and optimal distributions at different training stages on COCO, NUS, and AWA datasets under 5\% labeled setting, as shown in \figurename~\ref{fig:test_weight}. Here, the optimal distributions are computed using the true labels of the unlabeled data. The comparison on VOC is already presented in \figurename~\ref{fig1c}. As illustrated, our estimated distributions closely match the optimal correctness curves across datasets and training stages, demonstrating the reliability of our estimation strategy.
  \begin{table}[t]
    \footnotesize
    \setlength\tabcolsep{5pt}
    \centering
    \begin{tabular}{lccccl}
    \toprule
    \multirow{2}{*}{Weighting policy}  & \multicolumn{4}{c}{Labeled rate} & \multirow{2}{*}{Avg.}\\
      \cmidrule(r){2-5}
     & $5\%$ & $10\%$ & $15\%$ & $20\%$   & \\
    \midrule      
       Uniform       & $67.97$&$71.78$&$73.66$&$74.73$  & $72.04$\\
       Confidence       & $67.83$&$71.63$&$73.50$&$74.42$  & $71.84$\\
       Labeled      & $68.47$& $72.26$& $74.09$&$74.88$ & $72.42$\\
     \rowcolor{gray!10}  Ours  & $69.24$  &$72.86$&$74.44$&$75.48$  & $73.01$\\    
     \rowcolor{gray!30}  Optimal       & $69.32$&$72.93$& $74.57$&$75.54$  & $73.09$\\
 \bottomrule
    \end{tabular}%
    \caption{Effectiveness of weighting policies on COCO with various labeled rate. mAP(\%) is adopted as metric.}
    \label{tab:weight}%
  \end{table}
\subsubsection{Our weighting strategy is highly effective.}
\indent In Table~\ref{tab:weight}, we further evaluate the effectiveness of our estimated weights by comparing with several weighting strategies:
(1) \textbf{Uniform}: assigns all pseudo-labels a constant weight of 1;  
(2) \textbf{Confidence}: uses the predicted probability $p$ as the weight;  
(3) \textbf{Labeled}: estimates weights based on the labeled data;  
(4) \textbf{Optimal}: computes ideal weights based on ground-truth correctness on the unlabeled data. As shown in Table~\ref{tab:weight}, our proposed method (denoted as \textbf{Ours}) consistently outperforms all other weighting strategies. Notably, the \textbf{Confidence} strategy, which directly uses predicted probabilities, performs even worse than the \textbf{Uniform} baseline due to the unreliability of raw confidence scores. Meanwhile, our method achieves performance remarkably close to the \textbf{Optimal} setting, validating the high accuracy of our estimated correctness likelihoods and demonstrating the effectiveness of our weighting strategy.
  \begin{table}[t] 
    \footnotesize
    \setlength\tabcolsep{5pt}
    \centering
    \begin{tabular}{llcccc}
    \toprule
     Metrics & Method& VOC& COCO& NUS& AWA\\
    \midrule      
        \multirow{2}{*}{Time~(min)}&D2L& $1.85$&$21.42$&$37.74$&$8.38$
\\
        &Ours& $\mathbf{1.47}$&$\mathbf{18.29}$&$\mathbf{29.38}$&$\mathbf{6.61}$
\\ \midrule    
        \multirow{2}{*}{Memory~(GB)}&D2L& $7.34$& $14.17$& $14.13$&$14.12$
\\
        &Ours& $\mathbf{2.98}$&$\mathbf{4.44}$&$\mathbf{4.43}$&$\mathbf{4.44}$
\\    
 \bottomrule
    \end{tabular}%
    \caption{Time/memory comparison between D2L and our method on single 24GB NVIDIA GeForce RTX 3090 GPU. ``Time" is the training time per epoch, including threshold updating and weight estimating, ``Memory" is the max GPU memory allocated during training phase.}
    \label{tab:time}%
  \end{table}
\subsubsection{Our method is more efficient.} 
\indent As shown in Table~\ref{tab:time}, we compare the time and GPU memory usage of our method with the best compared method D2L on different datasets under the 5\% labeled ratio setting. From results demonstrated in Table~\ref{tab:time}, D2L requires significantly more GPU memory and has slower training speed due to its patch-wise image processing strategy. In contrast, our method achieves much lower GPU memory usage and faster training speed. Considering both the efficiency gains and the superior classification performance reported in Table~\ref{tab:main-result}, our method demonstrates clear advantages in both effectiveness and computational cost.

\section{Conclusion}
  \indent In this paper, we propose DiCaP, a Distribution-Calibrated Pseudo-labeling framework for SSMLL. We theoretically derive the optimal pseudo-label weight as the posterior correctness likelihood conditioned on prediction confidence. Motivated by the observation that correctness distributions remain stable across varying labeled data, we introduce a bin-based estimation scheme by partitioning the labeled data to approximate this optimal weight. To further reduce the impact of ambiguous predictions, we propose a dual-thresholding strategy that separates pseudo-labels into confident and uncertain regions. Finally, we introduce a fine-tuning stage to fully exploit all annotated data. Extensive experiments on multiple benchmarks validate the effectiveness of our approach under different label sparsity conditions.

\bibliography{aaai2026}
\end{document}